\documentclass[a4paper,twoside]{article}

\usepackage{epsfig}
\usepackage{subcaption}
\usepackage{calc}
\usepackage{amssymb}
\usepackage{amstext}
\usepackage{amsmath}
\usepackage{amsthm}
\usepackage{multicol}
\usepackage{pslatex}
\usepackage{apalike}
\usepackage{algorithm2e}
\usepackage[bottom]{footmisc}
\usepackage{url}
\usepackage{booktabs}
\usepackage{multirow} 
\usepackage{adjustbox}
\usepackage{xcolor}
\usepackage{SCITEPRESS}     

\def\ElectAI{{\sf ElectAI}}

\begin{document}

\title{Classifying Human-Generated and AI-Generated Election Claims in Social Media}

\author{Anonymous}

\author{\authorname{Alphaeus Dmonte\sup{1}, Marcos Zampieri\sup{1}, Kevin Lybarger\sup{1}, Massimiliano Albanese\sup{1} and Genya Coulter\sup{2}}
\affiliation{\sup{1}Department of Information Sciences and Technology, George Mason University, USA \\
\sup{2}OSET Institute}
\email{\{admonte, mzampier, klybarge, malbanes\}@gmu.edu, genya@osetinstitute.org}
}


\abstract{
Politics is one of the most prevalent topics discussed on social media platforms, particularly during major election cycles, where users engage in conversations about candidates and electoral processes. Malicious actors may use this opportunity to disseminate misinformation to undermine trust in the electoral process. The emergence of Large Language Models (LLMs) exacerbates this issue by enabling malicious actors to generate misinformation at an unprecedented scale. Artificial intelligence (AI)-generated content is often indistinguishable from authentic user content, raising concerns about the integrity of information on social networks. In this paper, we present a novel taxonomy for characterizing election-related claims. This taxonomy provides an instrument for analyzing election-related claims, with granular categories related to jurisdiction, equipment, processes, and the nature of claims. We introduce \ElectAI, a novel benchmark dataset that consists of 9,900 tweets, each labeled as human- or AI-generated. For AI-generated tweets, the specific LLM variant that produced them is specified. We annotated a subset of 1,550 tweets using the proposed taxonomy to capture the characteristics of election-related claims. We explored the capabilities of LLMs in extracting the taxonomy attributes and trained various machine learning models using \ElectAI\ to distinguish between human- and AI-generated posts and identify the specific LLM variant.}

\onecolumn \maketitle \normalsize \setcounter{footnote}{0} \vfill

\section{Introduction}

The widespread use of social media has fundamentally changed political discourse with direct impact on elections. Social media enables candidates and political entities to directly communicate with the electorate through platforms, such as X (formerly Twitter), profoundly shaping engagement strategies. This direct communication channel facilitates the rapid propagation of claims about election processes, including claims of fraud, by politicians and their supporters. Claims of fraud that are intentionally false or not fully supported by credible evidence can negatively impact election processes and compromise the integrity of elections. As a result, elections in many countries, such as the United States (US), have become more contentious in recent years\footnote{https://www.pewresearch.org/short-reads/2020/03/04/far-more-americans-see-very-strong-partisan-conflicts-now-than-in-the-last-two-presidential-election-years/}. 

The emergence of Large Language Models (LLMs) has revolutionized content generation across various domains. LLMs are capable of generating fluent and syntactically correct text that is virtually indistinguishable from human-generated text. LLMs often generate factually incorrect statements and, in some cases, may hallucinate~\cite{huang2023survey}. For example, Galactica~\cite{taylor2022galactica}, an LLM developed by Meta to summarize research papers and generate Wikipedia articles, was found to generate biased and incorrect content~\cite{heaven2022meta}. ChatGPT was also found to produce biased and inaccurate outputs~\cite{wach2023dark}. Furthermore, malicious actors can leverage the power of LLMs to intentionally produce uninformed claims and misinformation to spread them on social media~\cite{zhou2023synthetic}. 

The convergence of advanced LLMs and the widespread use of social media creates the perfect conditions for the spread of election misinformation. In this paper, we introduce a comprehensive taxonomy to further our understanding of human- and AI-generated election claims in social media. The taxonomy features several dimensions that often characterize election claims, such as jurisdiction, election infrastructure, and the nature of claims. We use the taxonomy to annotate \ElectAI, the first benchmark dataset for election claim understanding. \ElectAI\ contains a mix of 9,900 human- and AI-generated tweets in English. 

We present several experiments on claim characterization and authorship identification using \ElectAI\ with the objective to explore the following research questions:

\begin{itemize}
    \item {\bf RQ1:} How well do state-of-the-art LLMs understand the various attributes of election claims? 
    \item {\bf RQ2:} How do humans and machines compare in their ability to distinguish between human- and AI-generated election claims in social media?
\end{itemize}

\noindent Our main contributions are as follows:

\begin{enumerate}
    \item We propose a novel taxonomy to characterize and understand election-related claims and the attributes associated with these claims.
    \item We present \ElectAI, the first benchmark dataset for election claims. The dataset comprises both human- and AI-generated tweets annotated with the proposed taxonomy. We make \ElectAI\ freely available to the research community.\footnote{\url{https://languagetechnologylab.github.io/ElectAI/}}
    \item We present an evaluation of four LLMs (Llama-2, Mistral, Falcon, and Flan-T5) with respect to claim characterization on the \ElectAI\ dataset. 
    \item We evaluate the capability of multiple models and humans to discriminate between human- and AI-generated posts in the \ElectAI\ dataset. 
\end{enumerate}

The remainder of the paper is organized as follows. Section~\ref{sec:background} discusses the importance of fair and secure elections, motivating our choice of US elections as a case study, whereas Section~\ref{sec:related} discusses related work. Then, Section~\ref{sec:taxonomy} introduces the proposed taxonomy and Section~\ref{sec:electai} presents the \ElectAI\ dataset. Next, Sections~\ref{sec:understanding} and \ref{sec:attribution} present our evaluation of claim understanding and authorship attribution, respectively. Finally, Section~\ref{sec:conclusions} provides some concluding remarks and future research directions. 

\section{Background and Motivation} \label{sec:background}

Fair and secure elections are the backbone of democracy, and the spread of misinformation on social media can undermine their integrity. In 2017, former US Secretary of the Department of Homeland Security (DHS), Jeh Johnson, declared that \textit{``election infrastructure in this country should be designated as a subsector of the existing Government Facilities critical infrastructure sector. Given the vital role elections play in this country, certain systems and assets of election infrastructure meet the definition of critical infrastructure, in fact and in law.''}\footnote{Department of Homeland Security (2017). Statement by Secretary Jeh Johnson on the Designation of Election Infrastructure as a Critical Infrastructure Subsector. 
} As a result, US election infrastructure is considered critical infrastructure, as much as energy, transportation, or other systems critical to national security. Threats to election security can erode voter trust, delegitimize elections, and weaken democratic institutions. The spread of misinformation, disinformation, and malinformation (MDM) is a growing threat to fair and secure elections. MDM campaigns can target individual political candidates, local or state administrative entities, or even manufacturers of voting equipment. MDM campaigns erode trust in election processes and can also have significant financial consequences.

While the proposed approach has been demonstrated in the context of US elections, it can be easily generalized to create and study election-related datasets in any language or geopolitical context. 
American elections provide an interesting case study as they are inherently different from almost any type of election in other countries. With nearly 10,000 unique election jurisdictions, no other country has the level of decentralization that US elections do, and no other nation cedes as much control of federal elections to state and local bodies. As of 2020, over 260 million Americans were registered to vote. In jurisdictions such as Los Angeles County, California, voters will vote for more contests and candidates in one general election than the average voter overseas might vote for in their lifetime. 

As elections are conducted at the local level, there is great variability in the infrastructure and technical expertise of counties and towns within states. While the decentralization of elections avoids single points of failure, it requires each jurisdiction to manage and secure an extremely complex information technology system that includes voter registration databases, electronic poll books, vote-capture devices, optical scanners, vote tallying devices, election night reporting systems, and network technologies to securely transfer data. The complexity and diversity of this infrastructure further compound the challenges in assessing election-related claims. Thus, the US is the ideal test case for \ElectAI.

Another dimension contributing to the uniqueness of US elections is the emphasis on freedom of expression, allowing anyone to say virtually anything in any public forum. The United States of America is the world’s longest lasting representative democracy, and has served as a model to countless other nations inspired by the uniquely American ideals of independence, self-determination, and free expression.  For voters who live in the United States, there are very few restrictions on political speech by design. The First Amendment to the U.S. Constitution enshrines strong historic protections of freedom of expression, whether it is speech by individuals or by the press, free practice of religion, the right of assembly, as well as the right to petition the government for a redress of grievances. 

At the nexus of these widely admired but difficult to sustain ideals, is the social media platform now known as X, formerly Twitter, nimbly intersecting First Amendment rights and protections. While not as large as competing platforms such as TikTok or Facebook, it is fair to postulate that the cumulative impact of documenting current geopolitical events in real time with comparatively few barriers to entry has permanently enshrined X into the American lexicon. Elections, and the manner in which information about them is disseminated via social media channels, have been fundamentally transformed by the X platform. More Americans get their election information via X than they do from previously trusted sources such as local newspapers or from political parties. X offers an unprecedented level of instantaneous access to a voter’s local election officials, as well as direct communication with state and federal authorities. Failing to quantify the impact of AI-generated election misinformation on social media users now will set media literacy, public confidence in elections, and potentially, the stability of democratic processes themselves in imminent danger. By design, \ElectAI\ allows for a narrowing or broadening of scope, without limiting its application to just one field of research.

\section{Related Work} \label{sec:related}

There have been several studies addressing automatic verification of claims, rumor detection, and several other tasks related to the detection and mitigation of misinformation. Chen \emph{et al}.~\cite{chen2022gere} proposed a 3-stage pipeline for claim verification that (i) gets the encodings for the claims, (ii) retrieves relevant documents that main contain a support claim, and (iii) finds appropriate evidence in them. Yang \emph{et al}.~\cite{yang2022weakly} propose a Multi-Instance Learning (MIL) approach for rumor and stance detection. Barbera \emph{et al}.~\cite{barbera2018explaining} explored misinformation spread and the factors that contribute to it in the context of the 2016 US Presidential elections.

Recent developments in fact verification systems have significantly improved the accuracy and efficiency of automatic claim verification in various settings. Hu \emph{et al}.~\cite{hu2023read} proposed a fact-verification approach based on training an evidence retriever and a claim verifier with the retrieved evidence. The approach considers the faithfulness (the model's decision-making process) and plausibility (convincing to humans) of the retrieved evidence. While the great majority of studies on fact verification are on English texts, an approach for Arabic claim verification in social media was proposed by~\cite{sheikh2023tahaqqaq}. In the proposed pipeline, they first identify a claim presented in a post and then evaluate the trustworthiness of the claim. Multi-modal fact-checking has also been explored in a few studies such as those by Yao \emph{et al}.~\cite{yao2023end} and by Sun \emph{et al}.~\cite{sun2023med}, who introduced one of the datasets described next.  


There have been various datasets developed for claim understanding and verification. One of the most widely-used datasets is FEVER~\cite{thorne2018fever} developed for fact extraction and verification. Another dataset developed by Sun \emph{et al}.~\cite{sun2023med} targets multi-modal misinformation in the medical domain. The dataset contains multi-modal instances from news and tweets as well as instances generated by LLMs. A recent study by \cite{zhou2023synthetic} targets COVID-19 misinformation. The authors generate COVID-19-related misinformation using GPT-3 and they analyze the linguistic features of human- and AI-generated texts. The analysis indicates significant differences between the human and AI-generated text concerning the writing style, emotional tone, use of specific keywords, use of informal language, and several other features.

To the best of our knowledge, no other dataset has been created to address election claim understanding thus far. Furthermore, only a few datasets containing human- and AI-generated content have been curated for claim understanding. This motivates us to introduce \ElectAI\ and make it freely available to the research community. 

\section{Taxonomy} \label{sec:taxonomy}

\begin{figure*}[!ht]
    \centering
    \includegraphics[scale=.95]{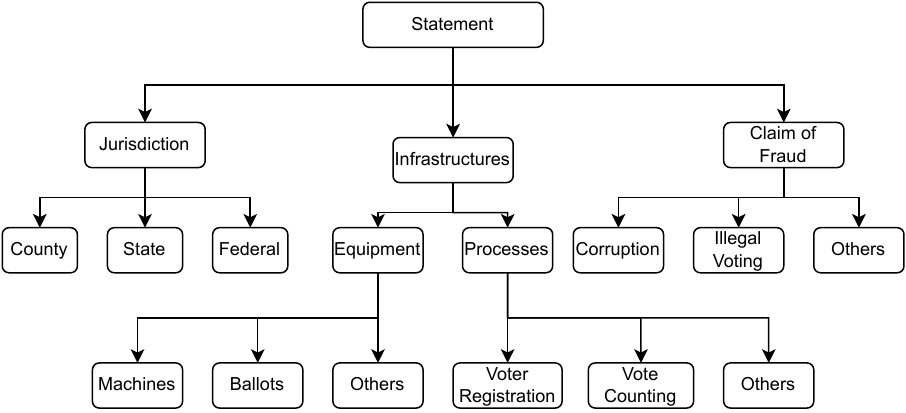}
    \caption{The election claim taxonomy.}
    \label{fig:taxonomy}
\end{figure*}

The proposed taxonomy, shown in Figure~\ref{fig:taxonomy}, was developed by identifying the most common set of attributes characterizing social media discourse about elections, and it was validated by subject matter experts in election administration. Two of the authors are regular attendees of an annual US-based election-focused conference bringing together academia, government, industry, and non-profit sector, and one author is a social media expert affiliated with a nonpartisan election technology research organization working to increase confidence in elections. Details are omitted to preserve the anonymity of the submission. 

The proposed taxonomy intentionally employs unambiguous naming conventions, with classification structures that can be commonly understood by a diverse group of academic researchers and election stakeholders, as well as applied to elections globally. While identical terms may not be employed worldwide, the foundational concepts of election jurisdiction, equipment, processes, and claims are commonly used by the election community globally and are highly relevant topics to the rapidly evolving field of AI and social media research. The proposed taxonomy will assist in the creation of a working set of comparative standards for future AI research endeavors while facilitating the efficient collection of quantifiable, high-quality training data for future machine learning initiatives. The race to develop data-driven methods for protecting election administrators and the voting public from the increasingly dangerous aftereffects of social media disinformation (particularly MDM generated by AI models) will be critical during the 2024 Presidential Election, when unprecedented amounts of malicious election-related social media content pose the risk of derailing American elections via ideological capture.

\subsection{Jurisdiction}

The Jurisdiction attribute refers to the government entity conducting the election and includes three subcategories:

\begin{itemize}
    \item \textbf{County}: Refers to a specific county, such as Fairfax County, Los Angeles County, or Monroe County.
    
    \item \textbf{State}: Refers to a specific state, such as New York (NY), Virginia (VA), or California (CA).
    
    \item \textbf{Federal}: Refers to federal elections, namely presidential elections.
\end{itemize}

\subsection{Infrastructure}

The Infrastructure attribute encompasses an array of components employed to carry out elections and includes the Equipment and Processes subcategories.

\noindent \textbf{Equipment}. Equipment refers to any system that is used to cast or count votes and includes three subcategories:
\begin{itemize}
    \item \textbf{Machines}: Refers to electronic voting equipment, such as direct recording electronic systems (DREs), ballot marking devices (BMDs), scanners, and electronic poll books (e-poll books).

    \item \textbf{Ballots}: Refers to paper ballots, including mail-in ballots.

    \item \textbf{Other}: Refers to any other type of voting equipment not captured by the Machines or Ballots subcategories.
    
\end{itemize}

\noindent \textbf{Processes}. Processes refers to activity conducted during the administration of elections and includes three subcategories:

\begin{itemize}
    \item \textbf{Voter Registration}: Refers to the process of registering eligible voters in a centrally managed voter registration database.

    \item \textbf{Vote Counting}: Refers to the process of tallying and tabulating votes cast in an election. This includes verifying the authenticity of ballots and systematically counting the votes to determine the outcome of the election. 

    \item \textbf{Other}: Refers to any other election process not captured by the Voter Registration or Vote Counting subcategories.
\end{itemize}

\subsection{Claim of Fraud}

The Claim of Fraud attribute refers to claims of election-related fraud in a media post and includes three subcategories:

\begin{itemize}
    \item \textbf{Corruption}: Refers to claims of corruption (e.g., a candidate offering money or favors in exchange for votes).

    \item \textbf{Illegal Voting}: Refers to claims of illegal voting (e.g., votes cast by illegal aliens or individuals voting multiple times).

    \item \textbf{Other}: Refers to any other claims of fraud not captured by the Corruption or Illegal Voting subcategories (e.g., ballot tampering or coercion tactics).

\end{itemize}


\section{The \ElectAI\ Dataset} \label{sec:electai}

\subsection{Data Collection and Generation}\label{sec:data-collection}


The \ElectAI\ dataset comprises a total of 9,900 tweets that were either human- or AI-generated. All tweets include a label indicating whether they are created by a human or AI, and all AI-generated tweets include a label indicating the LLM variant that was used to create them. A subset of 1,550 tweets were annotated using the proposed taxonomy, as reported in Table~\ref{tab:dataset-summary}, including 850 human-generated and 700 AI-generated tweets.

\begin{table}[ht]
    \centering \small
        \begin{tabular}{lrrr}
        \toprule
                        & \textbf{Human} & \textbf{AI} & \textbf{Total}\\
        \midrule
        \textbf{First Annotation Round} & 600 &  0 & 600\\
        \textbf{Second Annotation Round} & 250  & 700 & 950\\
        \bottomrule
        \textbf{Total} & 850  & 700 & 1,550\\
        \hline
    \end{tabular}
    \caption{Summary of annotated posts} \label{tab:dataset-summary}
\end{table}

To populate the human-generated part of the dataset, we initially sampled the existing VoterFraud2020 dataset~\cite{abilov2021voterfraud2020}, a multimodal dataset consisting of posts about potential events during the 2020 US presidential elections. Since the VoterFraud2020 dataset also includes tweets that are general statements, we use specific keywords like jurisdiction, equipment, process, device, scanner, counting, registration, corruption, etc. to identify and sample 850 tweets relevant to our task. We use the proposed taxonomy to create narrative frames, as described in recent work on COVID-19~\cite{zhou2023synthetic}, to generate 700 synthetic tweets using LLMs. These tweets were generated after the first round of annotations. The taxonomy attributes were used to generate tweets that spanned a diverse range of election-related topics and mimicked human-generated tweets. We use the following LLMs to generate the tweets.

\paragraph{\textbf{Falcon-7B-Instruct}} \cite{refinedweb} henceforth Falcon, is a decoder-only model fine-tuned with instruct and chat datasets. This model was adapted from the GPT-3 \cite{brown2020language} model, with differences in the positional embeddings, attention, and decoder-block components. The base model Falcon-7B, on which this model was fine-tuned, outperforms other open-source LLM models like MPT-7B and RedPajama, among others. The limitation of this model is that it was mostly trained on English data, and hence, it does not perform well in other languages.

\paragraph{\textbf{Llama-2-7B-Chat}} \cite{touvron2023llama2} henceforth Llama-2, is an auto-regressive language model with optimized transformer architecture. This model was optimized for dialogue use cases. The model was trained using publicly available online data. The model outperforms most other open-source chat models and has a performance similar to models like ChatGPT. This model, however, works best only for English.

\paragraph{\textbf{Mistral-7B-Instruct-v0.2}} \cite{jiang2023mistral} henceforth Mistral, is a transformer-based large language model which uses two architectures coupled together, a grouped-query attention along with sliding-window attention. The model uses a byte-fallback tokenizer. The model outperforms several open-source LLMs including Llama-2-13B on various NLP tasks.

The following prompt was used to generate the tweets.

\vspace{3mm}
\noindent\fbox{%
    \parbox{0.96\linewidth}{%
        \textit{\textbf{Question}: Write \textbf{\{\# of tweets\}} unique tweets using informal language and without the use of opinion statements, declarative statements, and call-to-action statements, about \textbf{\{claim of fraud\}} caused by \textbf{\{infrastructure\}} in \textbf{\{state\}}. You may choose the actual county in the state mentioned. You may include actual websites, people's names, and usernames. \\
        \textbf{Answer}:}%
    }
}
\vspace{3mm}

The prompt includes the \emph{claim of fraud}, \emph{infrastructure}, and \emph{state} attributes of the taxonomy. The labels from the taxonomy were substituted for prompt attributes, such as equipment and processes instead of infrastructure, corruption, and illegal voting instead of the claim of fraud, and mentioning states like Arizona, Virginia, etc. We only use the state-level jurisdictions as this better mimics the human tweets. However, the prompt includes instructions to include the county jurisdiction for some tweets. The prompt also includes specific instructions regarding the number of tweets to be generated. We use this dataset consisting of 1,550 human- and AI-generated tweets as an evaluation dataset for the claim understanding task, where we extract the claim-related attributes from the tweets.

In addition to exploring the extraction of claim-related attributes from human- and AI-generated tweets, we explored the ability of humans and machine learning models to distinguish between human- and AI-generated tweets. To facilitate this exploration, we generate additional tweets with Llama-2, Falcon, and Mistral using the prompt described above. For the human-generated portion, we sample the VoterFraud2020 dataset. Our training dataset for this authorship attribution task consists of 8,000 tweets and each tweet is labeled with \emph{human}, \emph{llama}, \emph{falcon}, or \emph{mistral} label, depending on the model used to generate the tweet. Table \ref{tab:example-train} shows example instances from the training dataset. 

\begin{table}[!ht]
    \centering \small
\resizebox{\columnwidth}{!}{%
\begin{tabular}{p{6.5cm}|c}
    \toprule
         \textbf{Tweet} & \textbf{Label}  \\
         \midrule
              @RealNews: Why are people trying to vote twice in Arkansas? Isn't that illegal? \#VoterFraud \#BaldFacedLie & human \\ \\
              Breaking news out of Guilford County! It looks like someone has been tampering with ballots! Stay vigilant and follow @integritywatchdog for updates. \#GuilfordCounty \#ElectionCorruption \#BallotTampering & llama \\ \\
              Voting rights in \#Iowa are in jeopardy! Anyone with information about voter registration irregularities or suspicious activity should contact @Iowa\_Election immediately - @InformedIowan & falcon \\ \\
              You know what's not cool? Voter registration irregularities in Kent County, Rhode Island. Hopefully the Board of Elections can sort this out before it's too late! \#CleanVoterRolls  & mistral \\
                    
        \bottomrule
    \end{tabular}
    }
    \caption{Example tweets from the training dataset.}
    \label{tab:example-train}
\end{table}

The claim characterization dataset is reused as a test dataset for the authorship attribution task. However, for this task, we re-annotate each tweet of the test dataset with a single label, depending on the model used to generate the tweet. We generate an additional 350 tweets using the Mistral model to be included in the test dataset.\footnote{These tweets are not annotated using the taxonomy described above as the Mistral model was not publicly available at the time of annotations. However, we included these these tweets in the authorship attribution task to have instances with all the labels, we include these these tweets.}

\subsection{Annotation}
We first developed a preliminary taxonomy that captured information related to jurisdiction, infrastructure, processes, and claims of fraud, similar to the final taxonomy presented in Figure~\ref{fig:taxonomy}. The preliminary taxonomy had more granular attributes (36 leaf nodes) than the taxonomy presented in Figure~\ref{fig:taxonomy} (12 leaf nodes). Using this preliminary taxonomy, a team of undergraduate and graduate students annotated 600 of the 850 human-generated tweets sampled from the VoterFraud2020 dataset. Each student annotated a sample of 50 tweets. Each tweet was annotated by up to three annotators, and a majority vote was used to adjudicate discrepancies and create the gold reference standard. We used this initial annotation effort to refine and consolidate the preliminary taxonomy to create the taxonomy shown in Figure~\ref{fig:taxonomy}.

Table~\ref{tab:annotation-scores} presents the inter-annotator agreement (IAA) for the taxonomy attributes, as well as the overall score. 

\begin{table}[!hb]
    \centering
     \scalebox{.95}{
    \begin{tabular}{l c}
    \toprule
         \textbf{Label}                             & \textbf{F1}  \\
         \midrule
        \textbf{Jurisdiction - State}               & 0.86 \\
        \textbf{Jurisdiction - County}              & 0.66 \\
        \textbf{Jurisdiction - Federal}             & 0.28 \\
        \textbf{Equipment - Machines}               & 0.79 \\
        \textbf{Equipment - Ballots}                & 0.70 \\
        Equipment - Other                           & 0.04 \\
        Processes - Voter Registration              & 0.16 \\
        \textbf{Processes - Vote Counting}          & 0.53 \\
        Processes - Other                           & 0.21 \\
        \textbf{Claim of Fraud - Corruption}        & 0.68 \\
        \textbf{Claim of Fraud - Illegal Voting}    & 0.65 \\
        Claim of Fraud - Other                      & 0.45 \\ \midrule
        \textbf{Overall}                            & \textbf{0.59} \\
                    
        \bottomrule
    \end{tabular}
    }
    \caption{IAA, measured as F1. Classes included in the claim understanding experiments are presented in bold.}
    \label{tab:annotation-scores}
\end{table}

\begin{figure*}[!ht]
    \centering
    \includegraphics[width=0.7\textwidth]{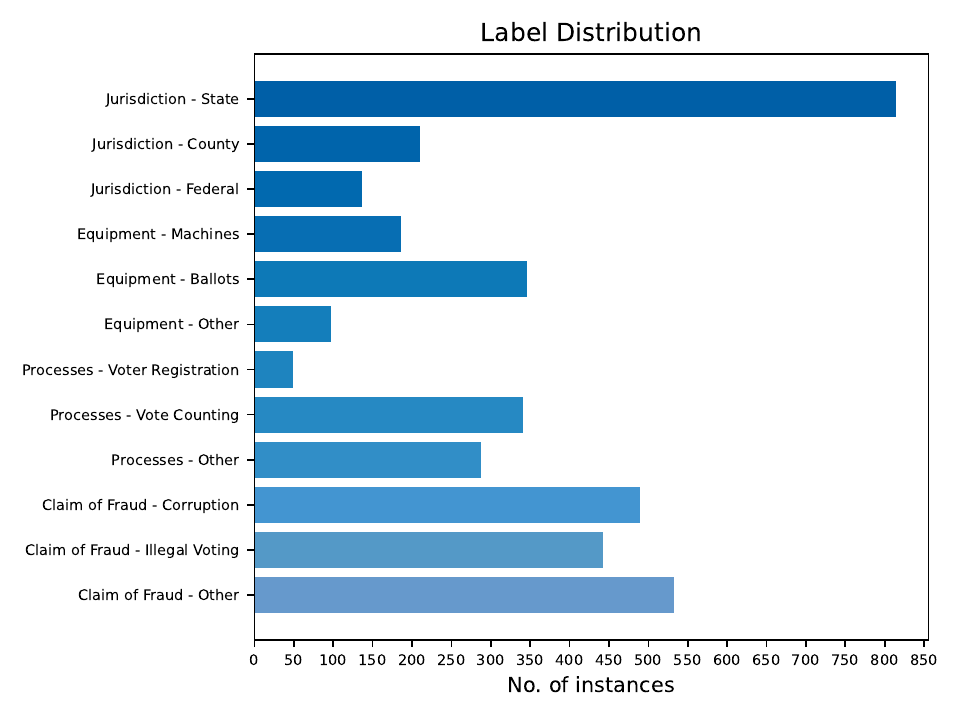}
    \caption{Statistics of the annotations. These statistics are based on the two rounds of annotation.}
    \label{fig:comb-stats}
\end{figure*}

We calculate IAA for the doubly annotated samples using F1-score, by holding one annotator as the reference and the other annotator as the prediction. We use F1 to provide a metric that can be compared with our classification results. We then used the revised taxonomy and guidelines to annotate an additional 250 human-generated tweets along with the 700 AI-generated ones, for a total of 950 tweets in round 2. Each tweet was annotated by up to three annotators. The data were annotated by undergraduate and graduate students in information technology and computer science. We provided detailed guidelines crafted based on the taxonomy, and the tweets were assigned a label using a majority vote.

As can be seen in Table~\ref{tab:annotation-scores}, there is a substantial to a high agreement for labels Jurisdiction -State, Jurisdiction - County, Equipment - Machines, Equipment - Ballots, Claim of Fraud - Corruption, and Claim of Fraud - Illegal Voting. However, the labels Jurisdiction - Federal, Equipment-Others, Processes - Voter Registration, Processes - Vote Counting, Processes - Other, and Claim of Fraud - Other, had fair to slight agreement. An overall F1-score of 0.59 indicates a moderate agreement between the annotators over all the labels. This level of agreement can be attributed to a low agreement for some labels.



\subsection{Data Statistics}

The 600 tweets from the first annotation effort were combined with the 950 tweets annotated using the refined taxonomy, to get a dataset with 1,550 annotated tweets, that we use for the claim understanding task. To maintain consistency between the annotations, we map the annotations from the first annotation effort to the refined taxonomy. Figure~\ref{fig:comb-stats} shows the statistics of the dataset for each label. As seen, most of the tweets had a mention of Jurisdiction-State. Whereas, not many tweets mention Processes-Voter Registration, Equipment-Others, and Jurisdiction-Federal. The mention of the various claims of fraud is almost evenly distributed between the tweets. We observed that some tweets mentioned more than one type of claim of fraud. We had a similar observation for other categories like Equipment, and Jurisdiction.

The authorship data set consists of tweets with multiclass labels associated with the tweet author, ${\emph{human}, \emph{llama}, \emph{falcon}, \emph{mistral}}$. Table \ref{tab:label-dist} shows the number of instances for each of the labels in the training and the test sets.

\begin{table}[!ht]
    \centering
\resizebox{\columnwidth}{!}{%
\begin{tabular}{c|c|c|c|c|c}
    \toprule
         \textbf{Dataset} & \textbf{llama} & \textbf{falcon} & \textbf{mistral} & \textbf{human} & \textbf{Total} \\
         \midrule
         Train & 2000 & 2000 & 2000 & 2000 & 8000 \\
         Test & 365 & 310 & 350 & 875 & 1900 \\
                    
        \bottomrule
    \end{tabular}
    }
    \caption{Label distribution for instances in the training and test datasets.}
    \label{tab:label-dist}
\end{table}

\section{Claim Characterization} \label{sec:understanding}

In this section, we report the performance of multiple LLMs in extracting the taxonomy attributes from the \ElectAI\ dataset.


\subsection{Large Language Models}\label{sec:llms}

Recently, LLMs have been widely used for several Natural Language Processing (NLP) tasks, including text generation and text classification. We use the three LLMs described in Section \ref{sec:data-collection} as well as the Flan-T5 model described below, for claim understanding (extraction of taxonomy attributes).




\paragraph{\textbf{Flan-T5-XL}} \cite{chung2022scaling} henceforth Flan-T5, is a language model based on the T5 \cite{raffel2020exploring} model, which is a Text-to-Text transformer model. This model was fine-tuned for better zero-shot and few-shot learning for over 1000 different tasks. The model is one of the few LLMs with support for languages other than English.

\subsection{Approach}

We approach the election claim understanding task as a question-answering (Q\&A) task, using LLMs in an in-context learning setting. In this Q\&A setting, the input prompt is a question focused on a single taxonomy attribute, along with the target tweet, and the output is a yes/no response, indicating whether the specific attribute is relevant to the tweet. The set of questions that span the taxonomy attributes is shown in Table \ref{tab:questions}. In initial experimentation, we found that the taxonomy attributes associated with Other, like Jurisdiction-Other,  were ambiguous and challenging for the models to extract. In mapping the taxonomy attributes to the questions in Table \ref{tab:questions}, we omitted the attributes associated with Other and instead included the associated parent label. For example, Jurisdiction-Other was replaced with a broader Jurisdiction label indicating whether any jurisdiction information is present (yes vs. no). We explored zero-shot learning, where no example inputs and outputs are provided in the prompt, and few-shot learning, where 3-5 input-output pairs are provided as examples in the prompt. However, the few-shot experimentation did not improve performance, so we only present the zero-shot results. Information extraction performance is evaluated using the F1-score. The following prompt was used to elicit model predictions.

\vspace{3mm}
\noindent\fbox{%
    \parbox{0.96\linewidth}{%
        \textit{\textbf{Tweet:} \textbf{\{tweet\}} \\
        Answer the following question with a yes or a no. \\
        \textbf{Question:} \textbf{\{question\}} \\
        \textbf{Answer:}
    }%
}
\noindent }
\vspace{3mm}


\begin{table}[!ht]
\normalsize
\centering
\resizebox{\columnwidth}{!}{%
\begin{tabular}{c|l}
    \toprule
         \textbf{ID} & \textbf{Question}  \\
         \midrule
              1 & Does the tweet mention a jurisdiction? \\
              2 & Does the tweet mention a state election? \\
              3 & Does the tweet mention a county election? \\
              4 & Does the tweet mention the federal election? \\
              5 & Does the tweet mention any election equipment? \\
              6 & Does the tweet mention electronic voting equipment machines? \\
              7 & Does the tweet mention ballots or related equipment? \\
              8 & Does the tweet mention any election-related process? \\
              9 & Does the tweet mention the vote counting process? \\
              10 & Does the tweet mention any election-related claim of fraud? \\
              11 & Does the tweet mention corruption in elections? \\
              12 & Does the tweet mention illegal voting? \\
        \bottomrule
    \end{tabular}}
    \caption{The input questions to the zero-shot prompts. Each question is based on an associated label from the taxonomy.}
    \label{tab:questions}
\end{table}


\subsection{Results and Discussion}

Table~\ref{tab:in-context} shows the F1 scores for the LLMs evaluated using the annotated benchmark dataset. 

\begin{table}[!ht]
\large
\centering
\resizebox{\columnwidth}{!}{%
\begin{tabular}{l|c|c|c|c}
    \toprule
         \multirow{2}{*}{\textbf{Label}} & \multicolumn{4}{c}{\textbf{F1}} \\
         \cmidrule{2-5}
         & \textbf{Llama-2} & \textbf{Mistral} & \textbf{falcon} & \textbf{Flan-T5} \\
         \midrule
              Jurisdiction & 0.828 & 0.851 & 0.554 & 0.898  \\
              Jurisdiction - State & 0.865 & 0.923 & 0.602 & 0.933  \\
              Jurisdiction - County & 0.562 & 0.852 & 0.705 & 0.824 \\
              Jurisdiction - Federal & 0.239 & 0.290 & 0.263 & 0.277 \\
              Equipment & 0.526 & 0.429 & 0.427 & 0.524 \\
              Equipment - Machines & 0.720 & 0.904 & 0.692 & 0.870 \\
              Equipment - Ballots & 0.510 & 0.609 & 0.540 & 0.549 \\
              Processes & 0.598 & 0.606 & 0.398 & 0.607 \\
              Processes - Vote Counting & 0.539 & 0.644 & 0.536 & 0.419 \\
              Claim of Fraud & 0.897 & 0.758 & 0.513 & 0.917 \\
              Claim of Fraud - Corruption & 0.452 & 0.523 & 0.497 & 0.544 \\
              Claim of Fraud - Illegal Voting & 0.577 & 0.611 & 0.317 & 0.571 \\
              \midrule
              Overall & 0.610 & 0.667 & 0.504 & 0.661 \\
                    
        \bottomrule
    \end{tabular}}
    \caption{Zero-shot in-context learning results. We report the F1 scores for individual labels as well as the macro-F1 score for each of the models}
    \label{tab:in-context}
\end{table}

Mistral achieved the best overall performance at 0.667 F1, followed by the Flan-T5 model at 0.661 F1. falcon achieved the lowest performance among the evaluated models at 0.504 F1. We also show the F1 scores for each of the labels used.
Apart from the falcon model, all the models had a better performance identifying the Jurisdiction and the Jurisdiction - State labels. Mistral and Flan-T5 models have a better performance for the Equipment - Machines label, as compared to the falcon and Llama-2 models. However, the performance of all the models in identifying the Jurisdiction-Federal label is low. Our analysis suggests that most tweets do not contain explicit references to federal elections and that federal jurisdiction is typically implicit. Performance for labels like Equipment and Processes is low. Some tweets may mention equipment, like voting machines, ballots, or other election equipment. However, the models struggle to identify this information from a given tweet. The tweets are annotated for particular labels even if some information is implicit. Hashtags are also considered during the annotation process. We observe that some LLMs are unable to identify this implicit information. For example, if the word ``ballots'' is preceded by a hashtag (``\#ballots''), the LLMs are unable to accurately identify the tweets referencing ballots. Another issue is the LLM's ability to understand and extract information from a given text. For example, the LLMs may label a tweet as mentioning a machine, even though voting machines are not mentioned in the tweet. Similarly, the LLMs may be unable to detect some specific terms and phrases like the name of a county if the word 'county' is not explicitly mentioned. These issues affect the performance of the LLMs, as evident from the results.

\section{Authorship Attribution} \label{sec:attribution}

In this section, we report on our evaluation of the performance of the Authorship Attribution Task, where the objective is to identify the author of the tweet, whether human or specific LLM.

\subsection{Classification Models}

We conducted experiments with various models as outlined below.

\paragraph{\textbf{Random Forest}}~\cite{breiman2001random} is a supervised machine-learning algorithm based on decision trees. It aggregates the outputs of multiple decision trees to produce the final output. Our experimentation involves utilizing two types of input features: Term Frequency - Inverse Document Frequency (TF-IDF) and Word2Vec 
embeddings~\cite{Mikolov2013EfficientEO,mikolov2013distributed,mikolov2013linguistic}, where we add the individual word vectors for each word in the sentence to output a sentence vector.

\paragraph{\textbf{Bidirectional Encoder Representations from Transformers (BERT)}}~\cite{devlin-etal-2019-bert} is a transformer-based model that has become ubiquitous in NLP. It has demonstrated state-of-the-art performance across various NLP tasks. In our study, we fine-tuned the BERT-based model with a multi-class classification objective. The BERT-base model is pre-trained on a substantial corpus of data, contributing to its robustness and effectiveness.

\paragraph{\textbf{Robustly Optimized BERT Pretraining Approach (RoBERTa)}} \cite{liu2019roberta} is a variation on BERT that uses a more dynamic masking strategy, which improves feature learning. Similar to BERT, we fine-tune this model for multi-class classification.

\subsection{Approach}

We conducted multi-class classification to distinguish between human- and AI-generated tweets and resolve the specific LLM author. Three models, as described previously, were either trained or fine-tuned for this task. The Random Forest classifier was trained using two sets of features: TF-IDF and Word2Vec embeddings. We utilized grid search to determine the optimal values for the two hyperparameters, \emph{N-estimators} and \emph{Max Depth}. Specifically, we explored values of N-estimators=[50,100,200] and Max Depth=[20,40,50]. For the BERT and RoBERTa models, we employed the AdamW optimizer with a learning rate of 1e-5. The models underwent training for 3 epochs. Subsequently, all models were fine-tuned using the training data and evaluated using the test dataset. We report the F1 score for each model.

\subsection{Turing Test}
We performed a Turing test to evaluate how well AI-generated tweets mimicked human-generated tweets. During the second phase of annotations, the annotators were asked to label the tweets as human- or AI-generated. These annotations were compared with the actual labels for the tweets. The overall accuracy in identifying the human- and AI-generated tweets was 36.5\%. We also calculated how accurately the annotators identified only the AI-generated tweets. We observed that only about 19\%  of the annotators were able to identify the AI-generated tweets. These results indicate that the AI-generated tweets are similar to human-written tweets. We observe specific patterns in the tweets generated by specific LLMs. Identifying these patterns and writing styles becomes straightforward if these tweets are presented consecutively. However, it may be difficult to identify these patterns if the tweets are presented randomly, especially with the human-written tweets included among the AI-generated tweets.

\subsection{Results and Discussion}


The best classifier for Random Forest with TF-IDF vectors had the N-estimators value of 200 with a Max Depth of 50, while the best classifier for Random Forest with Word2Vec Embeddings had N-estimators=200 and a Max Depth of 40. Both the BERT and the RoBERTa models achieved the maximum training F1 score at epoch 2.

\begin{table*}[!ht]
    \centering
    \scalebox{1}{
    \begin{tabular}{l|c|c|c|c|c}
    \toprule
         \multirow{2}{*}{\textbf{Model}} & \multicolumn{4}{c}{\textbf{Label}} \\
         \cmidrule{2-6}
         & \textbf{human} & \textbf{llama} & \textbf{falcon} & \textbf{mistral} & \textbf{Overall}\\
         \midrule
              Random Forest + TF-IDF & 0.982 & 0.638 & 0.752 & 0.861 & 0.837 \\
              Random Forest + Word2Vec Embeddings & 0.926 & 0.518 & 0.580 & 0.617 & 0.660 \\
              BERT & 0.991 & 0.964 & 0.975 & 0.953 & 0.971 \\ 
              RoBERTa & 0.998 & 0.974 & 0.987 & 0.975 & 0.983 \\
                    
        \bottomrule
    \end{tabular}}
    \caption{Model performance for the Authorship Attribution task. We report the F1 scores for each label as well as the macro-F1 scores for each of the models.}
    \label{tab:authorship}
\end{table*}

\begin{figure*}[!ht]
    \centering
        \begin{subfigure}{0.47\linewidth}
    		\includegraphics[width=\linewidth]{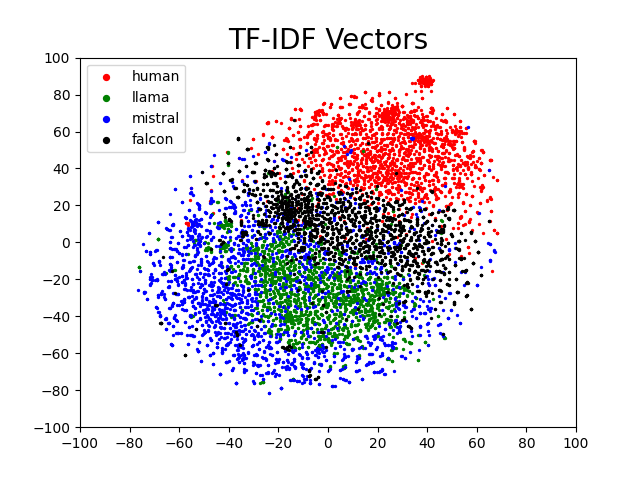}
    		\caption{TF-IDF Vectors}
    		\label{fig:tfidf-vec}
	  \end{subfigure}
	  \begin{subfigure}{0.47\linewidth}
    		\includegraphics[width=\linewidth]{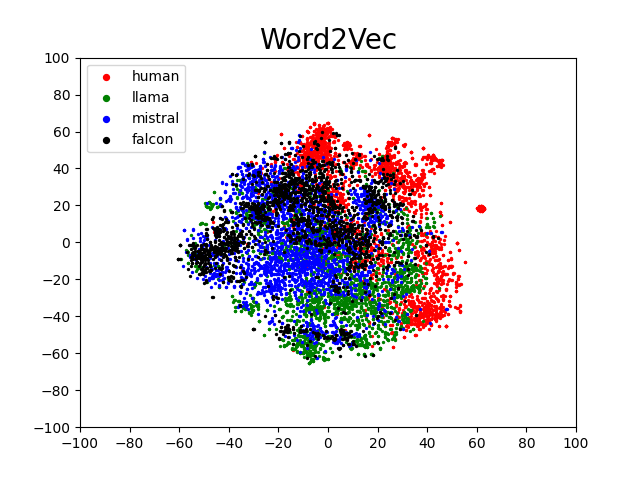}
    		\caption{Word2Vec Embeddings}
    		\label{fig:word2vec-emb}
	    \end{subfigure}
	\vfill
	    \begin{subfigure}{0.47\linewidth}
        		\includegraphics[width=\linewidth]{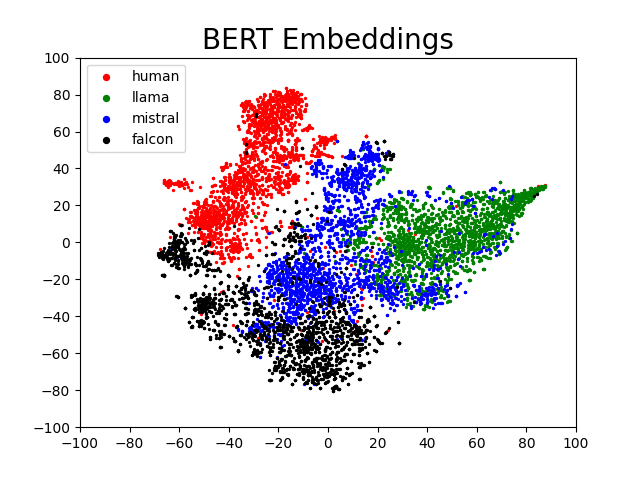}
        		\caption{BERT Embeddings}
        		\label{fig:bert-emb}
	    \end{subfigure}
	    \begin{subfigure}{0.47\linewidth}
    		\includegraphics[width=\linewidth]{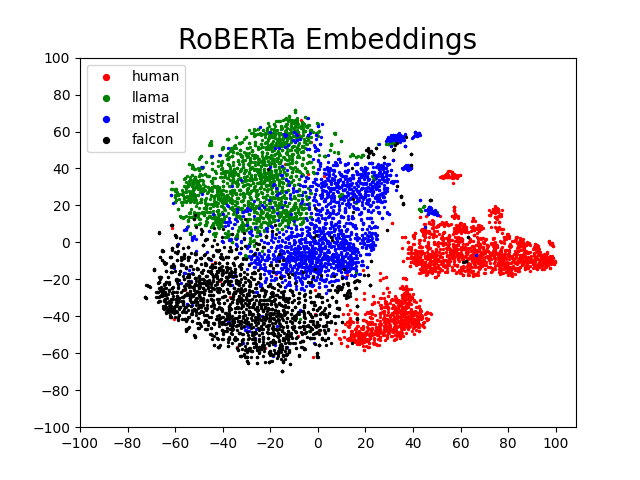}
    		\caption{RoBERTa Embeddings}
    		\label{fig:roberta-emb}
	    \end{subfigure}
    \caption{The clusters of the embeddings used to train the machine learning models for authorship attribution task}
    \label{fig:enter-label}
\end{figure*}

Table~\ref{tab:authorship} shows the F1 scores for each of the individual labels as well as the overall model. The transformer-based models, BERT and RoBERTa, achieve higher performance than the Random Forest models overall. All models have a superior performance identifying the human-generated tweets achieving a higher F1-score as compared to the tweets generated using LLMs. Moreover, these high performance of models, especially the transformer-based models can be attributed to the similarity between the tweets generated using several LLMs. We observe that tweets generated by each LLM tend to be homogeneous in sentence structure, vocabulary, phrasing, and other linguistic features. This makes it easier for the machine learning models to classify the tweets for the authorship attribution task, as they are exposed to a large number of tweets from each LLM in training.

To better understand the linguistic similarities and differences across the human and AI-generated tweets, we clustered the tweets based on the embeddings used to train the models, as shown in Figure \ref{fig:enter-label}. For the transformer models, we concatenated the last four hidden layers of the pre-trained BERT and RoBERTa models to create sentence embeddings, based on prior work~\cite{devlin-etal-2019-bert}. Note that these embeddings were generated using the pre-trained models, not the fine-tuned versions from our experimentation. We use t-SNE~\cite{van2008visualizing} to convert the high-dimensional embeddings to two-dimensional vectors. These are then plotted based on their respective author labels. We observe that clusters are formed based on the LLMs used to generate the tweets and these clusters are separable for the transformer-based models. Similarly, for the TF-IDF embeddings, the human-generated tweets form a separate cluster and are separable from the others. However, Word2Vec embeddings form small clusters, where the authors are not as easily separated. The separability of the clusters using the TF-IDF, BERT, and RoBERTa embeddings demonstrates that each LLM uses a distinct \textit{voice}, which makes it relatively easy to resolve the tweet author. There are inherent patterns in the tweets, which are visually indistinguishable from humans, however, machine-learning models can identify these patterns and isolate the tweets. However, these patterns become increasingly apparent when humans are provided with a greater quantity of tweets generated by a specific LLM.

\section{Conclusion and Future Work} \label{sec:conclusions}

We presented a novel taxonomy developed to further our understanding of election claims in social media. We used this taxonomy to curate \ElectAI, the first benchmark dataset for election claim understanding containing tweets generated by humans and AI. Using \ElectAI, we performed various experiments on claim characterization -- answering {\bf RQ1} -- and authorship attribution of human- vs. AI-generated tweets -- answering {\bf RQ2}. 

With respect to {\bf RQ1}, our results showed that although LLMs have achieved state-of-the-art performance on several NLP tasks, these models have a moderate performance on claim understanding in an in-context learning setting. Among the models tested, we showed that Mistral performed best with an overall performance of 0.667 F1 score while falcon performed worst at 0.504 F1 score. With respect to {\bf RQ2}, we showed that humans perform very poorly in discriminating between human- and AI-generated content. The results of a Turing Test show that humans can correctly discriminate human- from AI-generated with just over 36\% accuracy. Computational models, on the other hand, perform very well on this task with a RoBERTa model achieving a 0.983 F1 score in identifying the authorship of tweets in a four-class experiment with tweets generated by humans and three different LLMs. 

As part of our future work, we plan to annotate more instances of the \ElectAI\ dataset using the proposed taxonomy and conduct experiments on a larger scale. We plan to evaluate various approaches that can potentially improve the performance of the models, such as few-shot learning, instruction fine-tuning, and chain-of-thought prompting. Finally, the work presented here is a first step toward identifying misinformation in election claims. We further plan to extend this work to identify misinformation and verify the claims using fact-verification approaches.

\bibliographystyle{apalike}
{\small
\bibliography{custom}}


\end{document}